\documentclass[final,5p,times,twocolumn]{elsarticle}

\usepackage{amssymb}
\usepackage{amsmath}
\usepackage{latexsym}

\usepackage{url}
\usepackage{xcolor}
\usepackage{booktabs}
\usepackage{hyperref}
\usepackage{graphicx}
\usepackage{pdflscape}
\usepackage{subcaption}
\usepackage{comment}
\usepackage{float}
\usepackage{afterpage}
\usepackage{enumitem}
\usepackage{import}
\usepackage{tikz}
\usepackage{lipsum}
\usepackage[most]{tcolorbox}
\usepackage{makecell}

\usepackage{multirow}

\journal{XXX}

\begin{document}

\begin{frontmatter}


\title{Development and evaluation of CADe systems in low-prevalence setting: The RARE25
challenge for early detection of Barrett’s neoplasia}%

\author[1]{Tim J.M. Jaspers\corref{cor1}}
\cortext[cor1]{Corresponding author} 
\ead{t.j.m.jaspers@tue.nl}

\author[1]{Francisco Caetano}
\author[1]{Cris H.B. Claessens}
\author[1]{Carolus H.J. Kusters}
\author[2]{Rixta A.H. van Eijck van Heslinga}
\author[2]{Floor Slooter}
\author[2]{Jacques J. Bergman}
\author[1]{Peter H.N. De~With}
\author[2]{Martijn R. Jong}
\author[2]{Albert J. de Groof}
\author[1]{Fons van der Sommen}

\fntext[fn1]{This manuscript represents version 1 of the RARE Challenge report. The final author list is currently being finalized and will be updated in subsequent versions.}

\address[1]{Department of Electrical Engineering, Architectures for Reliable Image Analysis Lab, Eindhoven University of Technology, Eindhoven, Netherlands}
\address[2]{Department of Gastroenterology and Hepatology, Amsterdam University Medical Centers, University of Amsterdam, Amsterdam, The Netherlands}

\begin{abstract}
Computer-aided detection~(CADe) of early neoplasia in Barrett’s esophagus is a prototypical low-prevalence surveillance problem in which clinically relevant findings are rare. While many CADe systems report strong performance on balanced or enriched datasets, their behavior under realistic prevalence conditions remains insufficiently characterized. The RARE25 challenge was designed to address this gap by providing a large-scale, prevalence-aware benchmark for Barrett’s esophagus neoplasia detection. RARE25 comprised a publicly released training dataset and a hidden test set reflecting the incidence during real-world surveillance. Participating teams were evaluated using operating-point–specific metrics that emphasize high sensitivity and explicitly account for prevalence, enabling performance assessment under clinically meaningful conditions. In total, 11 teams from 7 countries submitted methods employing diverse architectures, pretraining strategies, ensembling techniques, and calibration approaches. Although several methods achieved strong overall discriminative performance, positive predictive values remained low for most approaches, emphasizing the intrinsic difficulty of low-prevalence detection and the risk of overestimating clinical utility when prevalence is ignored. Beyond ranking submissions, RARE25 provides insights into methodological gaps in current CADe research. All submitted methods relied on fully supervised classification, despite the dominance of normal findings in the training data. The absence of prevalence-agnostic formulations, such as anomaly detection or one-class learning, suggests an underexplored direction that may be better aligned with real-world deployment. By releasing a large public dataset and establishing a rigorous, reproducible evaluation framework, RARE25 aims to catalyze the development of CADe systems that are robust to prevalence shift and suitable for clinical surveillance workflows.
\end{abstract}

\begin{keyword}
Low-prevalence \sep Endoscopy \sep RARE \sep Barrett's neoplasia
\end{keyword}

\end{frontmatter}
\section{Introduction}
Early detection of clinically relevant abnormalities in surveillance settings is inherently challenging. Clinicians must detect rare and often subtle visual cues while navigating large volumes of normal or benign findings, a scenario known to increase miss rates even among experienced observers. This phenomenon is closely related to the well-established \emph{low-prevalence effect}, in which rare targets are disproportionately overlooked during visual search tasks, and to broader perceptual limitations that emerge when distinguishing subtle anomalies from normal background patterns~\citep{Wolfe2007LowPrevalenceErrors, Rich2008LowPrevalence}. In clinical practice, such misses contribute significantly to diagnostic error, recognized as a major patient-safety concern across healthcare domains. The stakes are particularly high when early-stage disease is involved, as timely identification is strongly associated with improved outcomes and more effective therapeutic options. Together, these factors illustrate why detection of rare and subtle lesions represents a persistent challenge in routine surveillance workflows, and why technological support systems are increasingly sought to mitigate human limitations.

Computer-aided detection~(CADe) systems have the potential to mitigate these risks by acting as a second reader or a triage aid in surveillance workflows, highlighting suspicious regions, and reducing the chance that a rare abnormality is overlooked. However, developing and evaluating CADe for low-prevalence surveillance settings is a unique challenge~\citep{Mori2025AIRareDisease, Jong2026LargeTrials}. The main difficulty is not the overall scarcity of data, as large collections of normal surveillance exams are generally available, but rather the inherently small number of positive~(anomalous) cases. Consequently, algorithms are typically trained on datasets that are artificially balanced for convenience or optimization, producing models that do not reflect the true class distribution encountered in practice. Properly dealing with this so-called \emph{prevalence shift} is especially crucial in the model deployment stages~\citep{GODAU2025103504}. Proposed systems are often trained and evaluated on datasets that over-sample rare classes, and are rarely evaluated under the real, heavily skewed prevalence that they will face after deployment. The consequence is twofold: (1)~models tuned on balanced training/test sets may exhibit optimistic performance metrics that do not generalize to clinical prevalence, and (2)~deployment-time prevalence shift can lead to an unacceptable amount of false positive predictions that burden clinicians and may cause \emph{alarm fatigue} or, conversely, conservative models that miss true positives. Therefore, robust benchmarking that reflects the expected clinical prevalence is crucial, yet remains uncommon in practice.

To highlight and address these difficulties, the RARE25 challenge has been organized. RARE, Recognition of Abnormalities in low-pREvalence cancer, more specifically focuses on the early detection of neoplasia in patients with Barrett’s esophagus~(BE), a paradigmatic surveillance problem. BE is a metaplastic change of the distal esophageal lining that arises in the context of chronic gastroesophageal reflux and confers an increased risk for progression to esophageal adenocarcinoma. Patients with BE enter routine surveillance programs involving periodic endoscopic examinations to detect dysplasia or early carcinoma at a stage amenable to endoscopic therapy~\citep{Weusten2023BarrettESGE}. Early detection matters: endoscopic therapy for mucosal neoplasia achieves high rates of complete response and durable organ-preserving outcomes~\citep{PECH2014652}, whereas advanced disease carries substantially worse survival~\citep{Desai2022PostEndoscopyBarrett, ACS2025EsophagusSurvival}. Since the annual risk of progression from non-dysplastic BE to adenocarcinoma is low~(estimates are substantially below 1\%~per year~\citep{Hvid-Jensen}), the prevalence of visible early neoplasia within surveillance image sets is exceptionally low. This makes BE surveillance an instructive exemplar for other low-prevalence cancer surveillance tasks~\citep{Scholvinck2017DetectionBarrett}. 

The RARE25 challenge was designed to provide a realistic and reproducible benchmark for CADe in this low-prevalence surveillance context. For the challenge, we have released the largest publicly available Barrett’s esophagus image collection to date~(over 3,000~labeled images) and measured performance in a locked private test set comprising more than 25,000~images sampled to reflect expected real-world prevalence. This design forces participants to confront both training-time class scarcity and deployment-time prevalence shift, as well as to evaluate algorithmic trade-offs in clinically meaningful operating regimes. The challenge ran as part of the EndoVis/MICCAI challenges and attracted broad international participation. 

This report describes the resulting benchmark, summarizes the methods submitted by the 11~participating teams originating from 7~countries, and analyses methods to provide insight into the specific problem of CADe for rare abnormalities in surveillance workflows. Beyond presenting rankings and quantitative results, we examine similarities between high-performing teams, missed instances, and practical design choices~(data sampling, architecture, loss functions, and calibration) that influenced performance. Our goal is to offer both a rigorous benchmark and a set of actionable recommendations for the future development and evaluation of CADe systems intended for low-prevalence clinical surveillance.

\section{Related works}

\begin{figure*}
    \centering
    \includegraphics[width=\textwidth]{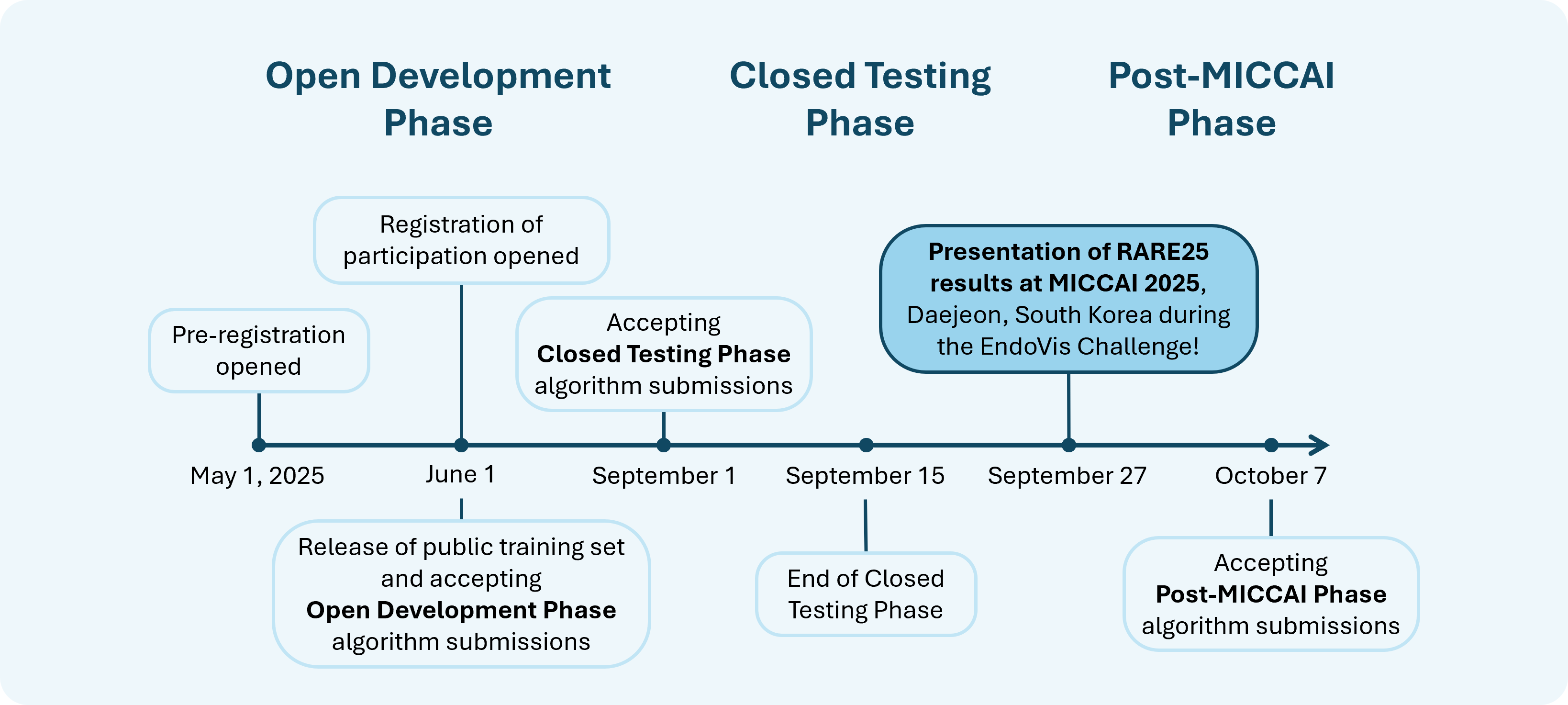}
    \caption{Overview of the RARE25 Challenge timeline, illustrating the three phases: Open Development Phase, Closed Testing Phase, and Post-MICCAI Phase.}
    \label{fig:timeline}
\end{figure*}

\subsection{Prevalence shift}

Recent work has highlighted prevalence shift, \emph{i.e.},~changes in the prior probability of disease between development and deployment settings, as a clinically important and distinct form of dataset shift that is insufficiently addressed by common model-development workflows~\citep{subbaswamy2020development}. \cite{zhang2022shifting} argue that a central challenge in translating machine learning for healthcare into clinical impact lies in the shift from controlled development environments to real-world deployment, where differences in data distributions, clinical practices, and sampling mechanisms can degrade model reliability if not explicitly addressed. Common causes of prevalence changes also include sample selection bias and variations in environmental factors such as season~\citep{van2019calibration, dockes2021preventing, zhang2022shifting}. According to \cite{dockes2021preventing} and \cite{GODAU2025103504}, handling prevalence shift is especially crucial in the following stages related to model deployment: (a)~model recalibration, (b)~decision rule~(the mapping of continuous predicted class scores into a single categorical decision), and (c)~performance assessment. Additionally, \cite{GODAU2025103504} introduces a new workflow for handling prevalence shifts that compensates for pitfalls. Lastly, \cite{Roschewitz2025} proposed a method to automatically detect data shifts, including prevalence shifts. 

\subsection{Low-prevalence benchmarks}

Low-prevalence anomaly detection is a recurring challenge across machine-learning domains, characterized by a pronounced imbalance between abundant normal data and scarce anomalous events. This setting has been extensively studied in industrial inspection, where benchmarks such as MVTec AD~\citep{bergmann2019mvtec} and Kaputt~\citep{hofer2025kaputt} adopt one-class or unsupervised anomaly detection protocols, training exclusively on defect-free samples. Related formulations appear in video surveillance, where benchmarks such as ShanghaiTech~\citep{luo2017revisit} and UCF-Crime~\citep{sultani2018real} address temporally sparse anomalies using either normal-only training or weakly supervised multiple instance learning. These works demonstrate that low-prevalence learning is not domain-specific but a broadly relevant problem with established methodological paradigms.

In the medical domain, however, low-prevalence detection introduces additional complexity due to substantial biological heterogeneity and stringent clinical requirements. Recent efforts, such as BMAD~(Benchmarks for Medical Anomaly Detection)~\citep{bao2024bmad}, standardize unsupervised anomaly detection across medical imaging modalities, often relying on healthy-only training splits, as seen in datasets such as Camelyon16~\citep{Camelyon2016}. Closely related challenges arise in dermatology, where the ISIC challenge datasets~\citep{gutman2016skinlesionanalysismelanoma} exhibit severe class imbalance, with malignant lesions representing only a small fraction of the available data. Unlike industrial or surveillance data, healthy tissue exhibits wide stochastic variability, forcing models to learn feature representations that capture a broad normal distribution while remaining sensitive to subtle pathological outliers. As a result, medical anomaly detection shifts the emphasis from balanced classification toward clinically realistic outlier detection, where sensitivity to rare, early-stage pathology and precise localization are paramount; a framing that directly aligns with the detection of early Barrett’s neoplasia.

\subsection{Metrics for low-prevalence model evaluation}
The evaluation of algorithms in highly imbalanced clinical settings is complicated by the fact that many commonly reported metrics are insensitive to the practical consequences of class imbalance. A recent multistage Delphi process identified the disregard of dataset properties, in particular, extreme class imbalance, as a frequent pitfall in medical-image evaluation~\citep{Reinke2024}. \cite{Hicks2022} also emphasized that special care should be taken in cases of imbalanced classes. In low-prevalence surveillance tasks, accuracy is dominated by the majority class and can therefore be misleading, while even balanced accuracy may overestimate clinical usefulness: although it weights sensitivity and specificity equally, a model with a low but non-negligible false-positive rate can still generate a prohibitive absolute number of false alarms when deployed under real-world prevalence. Similarly, threshold-free ranking metrics such as the Area Under the Receiver Operating Curve~(AUROC) and Area Under the Precision-Recall Curve~(AUPRC) capture complementary aspects of discrimination and retrieval performance but do not, by themselves, reflect the operational burden associated with false positives or the dependence of predictive values on prevalence. Recent analysis has shown that neither AUROC nor AUPRC should be treated as universally preferable under imbalance; instead, their interpretation depends on the clinical operating regime and decision context~\citep{McDermott2024}.

\subsection{CADe systems for Barrett's neoplasia}
Recent advances in computer vision, particularly driven by deep convolutional neural networks and vision transformers, have stimulated growing interest in CADe systems for Barrett’s esophagus surveillance~\citep{Ebigbo2019, DEGROOF2020, Meinikheim2024AIBarrett, FOCKENS2023, Jong2025CADeBarrett, KUSTERS2025108891}. Several studies have demonstrated that CADe systems can achieve expert-level performance under controlled experimental conditions, raising expectations that such systems may support clinicians during routine endoscopic surveillance by acting as a second observer~\citep{FOCKENS2023}.

Despite these advances, progress in the automated detection of Barrett’s neoplasia remains constrained by the limited availability of large, well-annotated public datasets. GastroVision, a large-scale endoscopic dataset covering the full gastrointestinal tract, includes a small subset of Barrett’s esophagus images~(95 images) but does not provide annotations indicating the presence or absence of neoplasia~\citep{GastroVision}. The EDD2020 challenge dataset does include Barrett’s esophagus cases with annotated neoplastic lesions; nevertheless, the complete dataset comprises only 384 images~\citep{EDD2020}. Earlier efforts, such as the EndoVis 2015 sub-challenge focusing on Barrett’s neoplasia detection, provided a highly targeted dataset, albeit limited in scale, comprising only 100 nondysplastic and 100 neoplastic images~\citep{EndoVIS_Barrett}. As a result, many CADe systems are trained and evaluated on either in-house private datasets or small public datasets, which limits reproducibility and cross-study comparability.

\subsection{Position of RARE25}
RARE25 occupies a complementary but distinct niche relative to prior work: rather than exploring early neoplasia detection under idealized or balanced conditions, the challenge is explicitly engineered to reproduce the deployment realities of surveillance imaging. By coupling a large labeled training set with a private test set sampled to reflect expected clinical prevalence, RARE25 forces teams to confront both training-time class scarcity and deployment-time prevalence shift, and to evaluate practical trade-offs that directly affect clinical workflow. Where the existing open source benchmarks either emphasize one-class/unsupervised anomaly protocols or small curated diagnostic datasets (prior Barrett’s collections), RARE25 prioritizes prevalence realism and operational relevance. Concretely, the challenge contributes (1)~a substantially larger public Barrett’s image collection than previously available, (2)~a locked test bed sampled to clinical prevalence to reveal realistic false-positive burden, and (3)~evaluation that highlights clinically meaningful operating points.

\section{Challenge organization}
The RARE25 challenge was organized as part of the EndoVis workshop at the MICCAI 2025 conference in Daejeon, South Korea. The challenge was jointly coordinated by the ARIA lab at Eindhoven University of Technology and the Amsterdam University Medical Centers, aligning with the broader goals of the BONSAI consortium to advance the automated detection of early neoplasia in Barrett’s esophagus. The challenge was hosted on the \texttt{Grand-Challenge.org} platform, where all phases, leaderboards, and submissions were managed. Fig.~\ref{fig:timeline} illustrates the key milestones and timeline of the challenge, from pre-registration through testing and results dissemination. 

Participation in RARE25 was open to the global research community. The challenge was structured into two main phases. The Open Development Phase began concurrent with the public release of the training data and lasted three months, during which participants could submit model outputs to the validation leaderboard to support iterative method development. A submission limit of one validated result per team per day was enforced to encourage meaningful improvements rather than overfitting based on the leaderboard. Following this, the Closed Testing Phase allowed each team one final submission for evaluation on the held-out test data, with results kept confidential until formal presentation at EndoVis/MICCAI 2025. Finally, the challenge reopened during the Post-MICCAI phase; however, no additional submissions were made on the final test set.

A total of 226 individual participants registered for the challenge, collectively representing multiple research groups and institutions worldwide. During the Open Development Phase, 196 submissions were received from 20 teams across 10 countries. These submissions reflect wide engagement with the low-prevalence classification task and highlight the challenge’s role in benchmarking real-world performance. Finally, 11 teams from 7 countries completed their submissions to the Closed Testing phase.
 
Participation policies were designed to ensure fairness and reproducibility. Only fully automated methods were permitted; participants were restricted to using the provided training data and publicly available datasets, while the use of publicly pretrained models was allowed. All contributing teams were invited to co-author this challenge report, and participants agreed not to publish challenge results prior to the release of this joint paper. 

\section{Challenge design}
\begin{figure*}
    \centering
    \includegraphics[width=\textwidth]{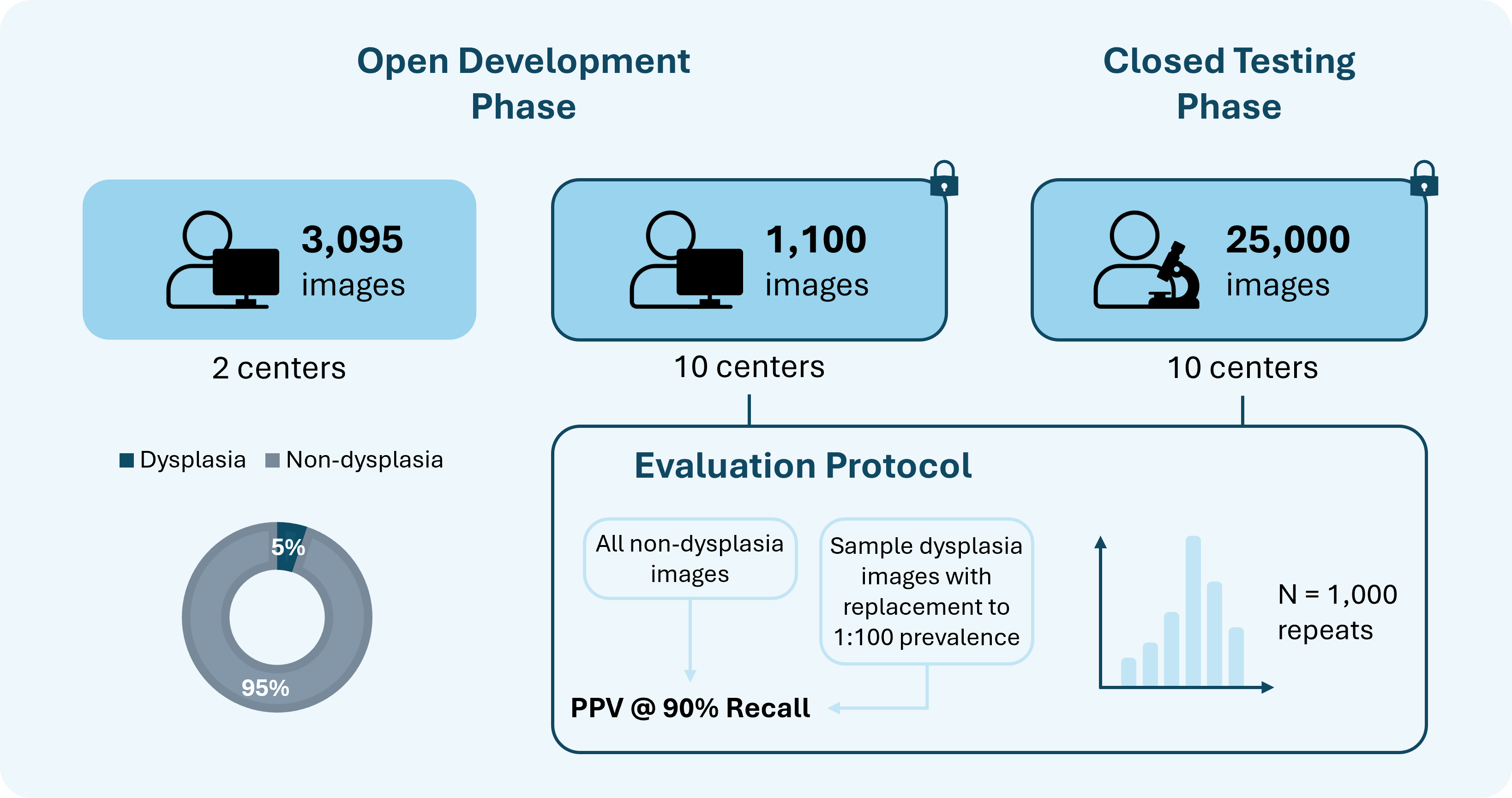}
    \caption{Overview of the RARE25 Challenge setup, highlighting the dataset class imbalance and the evaluation protocol.}
    \label{fig:setup}
\end{figure*}

Fig.~\ref{fig:setup} visually illustrates the overall challenge design, summarizing the class distribution in the public training data and the evaluation protocol for the closed testing phase.

\subsection{Public training set}
The public training set provided to all challenge participants comprises fully anonymized endoscopic images released under a non-commercial \texttt{CC-BY-NC-SA} license. The collection contains 2,937 images of non-dysplastic Barrett’s esophagus and 158 images of confirmed neoplasia~(3,095 images in total), making it, to our knowledge, the largest publicly available dataset of endoscopic images focusing on Barrett’s esophagus to date.

\begin{figure*}
    \centering
    \includegraphics[width=\textwidth]{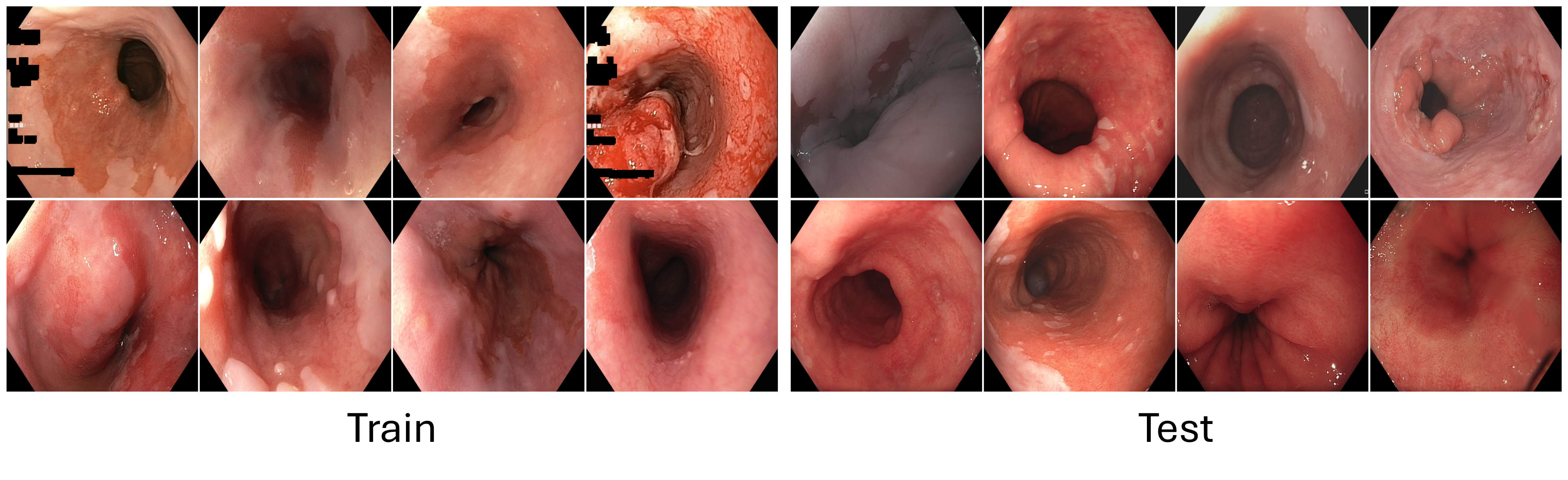}
    \caption{Visual examples of images included in the public training and closed test set.}
    \label{fig:data_examples}
\end{figure*}

All images in the training set were acquired using the Exera III endoscopy system~(Olympus Corp., Tokyo, Japan). Images were obtained retrospectively from the EndoBase archive via an automated query that retrieved stored examinations. From this pool, images related to Barrett’s esophagus were manually identified and extracted. No standardized imaging protocol governed the original acquisitions; consequently, the appearance of images and technical parameters~(illumination, zoom, white-balance, imaging mode, etc.) reflect routine clinical practice and therefore vary according to local workflows and operator preferences. The training data were collected at two Dutch medical centres during routine surveillance procedures; because collection occurred as part of standard care, detailed records of operator experience were not available for every case. All images were de-identified prior to release, which introduced compression-related anonymization artifacts in a subset of the training images~(Fig.~\ref{fig:data_examples}). These artifacts are absent from the test data.

\subsection{Private validation and test sets}
The challenge evaluation used closed~(private) validation and test cohorts that were retained by the organizers and not released publicly. The validation cohort comprises 1,040 non-dysplastic Barrett’s esophagus~(NDBE) images and 103 neoplasia~(NEO) images. The held-out testing cohort contains over 25,000 images~(23,176 NDBE, 3,232 NEO); both cohorts were sampled to reflect the low prevalence of Barrett’s neoplasia expected in routine surveillance.

All validation and test images were also acquired using the Exera III system. The cohorts were assembled using a combination of retrospective and prospective data collection. Retrospective images were retrieved from EndoBase using pre-identified patient lists linked to pathology reports; retrieved images were manually reviewed to determine the presence or absence of visible neoplastic lesions. Prospective images were collected according to a standardized acquisition protocol implemented by expert endoscopists specializing in Barrett’s surveillance. The validation and test cohorts include cases contributed by 12 medical centres participating in the BONSAI consortium~\citep{Fockens2023DeepLearningBarretts}.

\subsection{Evaluation Protocol}
Given the extreme class imbalance characteristic of Barrett’s esophagus surveillance, the performance evaluation was designed to be both prevalence-aware and statistically robust. In addition to computing metrics on the full held-out dataset, we employed image-level bootstrapping with 1,000 iterations to estimate uncertainty and confidence intervals. In each bootstrap iteration, all NDBE images were retained. In contrast, replacement neoplasia images were sampled to maintain a fixed 100:1 imbalance ratio (NDBE:NEO), reflecting the expected prevalence in real-world surveillance. This strategy preserves the dominant negative class while enabling stable estimation of metric variability in the rare positive class.

Following the proposed Metrics Reloaded report~\citep{MaierHein2024MetricsReloaded}, three complementary metrics were calculated: the area under the receiver operating characteristic curve~(AUROC), the area under the precision-recall curve~(AUPRC), and the positive predictive value at 90\% recall~(PPV@90RECALL). AUROC and AUPRC quantify global discriminative and retrieval performance, respectively, while PPV@90 captures performance at a clinically relevant operating point, explicitly reflecting the trade-off between recall and false-positive burden in low-prevalence surveillance. The reported final scores correspond to the median of the bootstrapped metric distributions, with 95\% confidence intervals derived from the 2.5th and 97.5th percentiles. For completeness, metrics computed on the complete test data set without resampling are also reported.

This evaluation protocol was chosen to balance ranking-based assessment, operating-point relevance, and statistical robustness under extreme class imbalance, and to provide insight into both expected performance and uncertainty when deploying CADe systems in real-world Barrett’s esophagus surveillance settings.

\section{Participating teams and methods}

Table~\ref{tab:teams} summarizes the teams that successfully submitted a method to the final test phase of the RARE25 challenge. In total, 11 teams from 7 countries completed a valid submission to the closed testing phase; no extra submissions were made during the Post-MICCAI phase. The table provides a high-level overview of the main methodological components reported for each final submission. The following subsections describe the individual approaches adopted by each team in more detail, while the final subsection presents a comparative analysis that synthesizes common methodological themes and design choices across submissions.

\begin{table*}[t]
    \scriptsize
    \setlength{\tabcolsep}{5pt}
    \centering
    \begin{tabular}{l | p{5.6cm} | c c c c c c}
        \toprule
        \textbf{Team} &
        \textbf{Affiliation} &
        \textbf{Architecture} &
        \textbf{Ens.} &
        \textbf{Add. data} &
        \textbf{Pretrain} &
        \textbf{IV }&
        \textbf{Calib.} \\
        \midrule

        IMSY &
        \makecell[l]{German Cancer Research Center (DKFZ),\\
        Division of Intelligent Medical Systems, Germany} &
        \makecell{ResNet-50\\ViT-L (DINOv3)} &
        \makecell{40-model\\ensemble} &
        No &
        \makecell{GastroNet-5M\\DINOv3 (LoRA)} &
        5-fold &
        Yes \\

        UT &
        \makecell[l]{The University of Tokyo,\\Japan} &
        MaxViT-T &
        No &
        No &
        ImageNet &
        5-fold &
        No \\

        Jmees-inc &
        \makecell[l]{Jmees-inc,\\Japan} &
        \makecell{ViT-B +\\ViT-L (seg.)} &
        \makecell{2-model\\ensemble} &
        Yes &
        \makecell{Segmentation\\pretraining} &
        5-fold &
        No \\

        MEDAI &
        \makecell[l]{Konkuk Univ.; Pukyong Natl. Univ.;\\
        Seoul St. Mary’s Hospital, South Korea} &
        ResNet-50 &
        No &
        \makecell{External\\val set} &
        GastroNet-5M &
        \makecell{External data\\\citep{EndoVIS_Barrett}} &
        Yes \\

        AJLG\_ETRI &
        \makecell[l]{Ajou University; ETRI,\\South Korea} &
        ConvNeXt-B &
        No &
        No &
        ImageNet &
        5-fold &
        No \\

        sk &
        \makecell[l]{Muroran Institute of Technology;\\
        Fujita Health University, Japan} &
        ConvNeXt-B &
        \makecell{10-model\\ensemble} &
        No &
        ImageNet-21k &
        50/50 &
        No \\

        agaldran &
        \makecell[l]{Tecnalia,\\Spain} &
        ResNet-50 &
        \makecell{25-model\\ensemble} &
        No &
        \makecell{GastroNet-5M} &
        5-fold &
        No \\

        SAINTS &
        \makecell[l]{University of York,\\United Kingdom} &
        DeiT-B &
        No &
        No &
        ImageNet-21k &
        80/20 &
        No \\

        ReMIC &
        \makecell[l]{OTH Regensburg,\\Germany} &
        \makecell{ConvNeXt\\+ Hiera} &
        \makecell{4-model\\ensemble} &
        Yes &
        ImageNet &
        80/20 &
        No \\

        MIRC &
        \makecell[l]{KU Leuven,\\Belgium} &
        ResNet-50 &
        No &
        No &
        ImageNet &
        Center-split &
        No \\

        iRARE &
        \makecell[l]{ICANS, ICUBE IMAGeS,\\
        University of Strasbourg, France} &
        ResNet-50 &
        No &
        No &
        GastroNet-5M &
        80/20 &
        No \\

        \bottomrule
    \end{tabular}
    \caption{Overview of participating teams and their final submissions. Abbreviations: IV = internal-validation split; Calib. = explicit post-hoc calibration.}
    \label{tab:teams}
\end{table*}

\subsection{Team IMSY}
Team IMSY tackled the challenge of limited training data and potential distribution shifts from unseen centres in the test set by leveraging large foundation models with diverse pretraining. Their approach combined a domain-specialist ResNet50 model pretrained on GastroNet-5M \citep{JONG2025} with a LoRA fine-tuned DINOv3 ViT-Large model~\citep{siméoni2025dinov3}. Additional external data were not used, ensuring a fair evaluation under the challenge constraints.

To improve generalization, the team applied a combination of standard augmentations—such as resizing, cropping, flipping, rotation, and color jittering, as well as targeted perturbations identified through center-to-center validation. These included geometric transformations, blur, color shifts, and optical distortions. The ensemble consisted of 40 models: 20 ResNet50 and 20 ViT-Large checkpoints derived from five-fold cross-validation, with each fold trained using four different augmentation presets. Predictions were probabilistically combined, and logits were post-processed with affine re-calibration using validation data to account for the expected 1:100 class imbalance~\cite{GODAU2025103504}.

For model optimization, DINOv3 ViT-Large models were fine-tuned with LoRA using cross-entropy loss with class weights and a balanced sampler, while ResNet50 models employed a surrogate loss targeting PPV at 90\% recall. AdamW optimizers were used for both architectures, with learning rates and weight decay selected via internal hyperparameter tuning. This training setup ensured robust convergence despite the small dataset.

In internal development, the team also explored alternative train/test splits to evaluate generalization across centres, including center-based splits and a separate internal five-fold cross-validation with a reserved test set. The input images were processed at a resolution of 224×224. Overall, the combination of domain-specialist models, foundation model fine-tuning, and ensemble recalibration allowed the team to effectively address data scarcity and distribution shifts.

\subsection{Team UT}
Team UT proposed a MaxViT-Tiny transformer-based model~\citep{MaxViT} to detect low-prevalence cancer abnormalities in medical images. MaxViT combines convolutional and attention mechanisms, making it well-suited for fine-grained feature extraction in medical imaging. The method was trained solely on the challenge dataset and used a 5-fold cross-validation stratified by group to prevent data leakage and ensure robust evaluation.

To improve generalization, the team applied a comprehensive set of augmentations during training, including geometric transformations, flips, shifts, scaling, rotations, and coarse dropout. Images were processed at a resolution of 384×384 pixels, normalized using ImageNet statistics, and training employed mixed precision with automatic gradient scaling to optimize memory usage and training efficiency. The binary classification task (normal vs. abnormal) was optimized using Binary Cross Entropy with Logits Loss and the Adam optimizer with cosine annealing warm restarts.

The model achieved strong cross-validation performance, with ROC-AUC scores ranging from 0.934 to 0.960 across folds (mean 0.9457, SD 0.0100). Training converged rapidly within five epochs, and validation losses stabilized between 0.06 and 0.10. However, domain shift challenges were significant: when evaluated on a separate development set, performance dropped to ROC-AUC 0.7709 and PPV@90\%RECALL 0.0112, highlighting the difficulty of generalizing across imaging conditions and centres. The team also experimented with medical-domain pretraining using GastroNet models~\citep{BOERS2024103298} to address domain shift, but these achieved lower performance than ImageNet-pretrained MaxViT models. 

\subsection{Team Jmees-inc}
Team Jmees-inc reframed the challenge from image-level classification to a segmentation-driven approach, motivated by the difficulty of distinguishing early neoplasia from non-dysplastic Barrett’s esophagus when spatial cues are not available. By training models to localize suspicious regions, the method aims to enhance interpretability and improve discrimination between subtle lesion appearances. Their strategy relied on extensive staged pretraining: starting with large polyp segmentation datasets from the same endoscopy domain, progressing to Barrett’s esophagus segmentation using EDD2020 and EVC datasets, and finally transitioning to cancer-specific segmentation and classification tasks. The learned representations were then fine-tuned on RARE25 using classification only.

The final system combined two transformer-based models: (1) a DINOv3-pretrained ViT-Base with a DPT segmentation decoder and (2) a ViT-Large model pretrained using masked image modeling. During fine-tuning on RARE25, only the classification heads remained trainable, with the segmentation components frozen. Training involved a set of standard endoscopic augmentations, high-resolution inputs (512×512 and 384×384), and exponential moving average updates for stable optimization. The predictions of the two models were averaged to produce the final ensemble output.

Using stratified 5-fold cross-validation, the approach achieved strong results, including AUROC 0.9896, AUPRC 0.7408, and PPV 0.235, indicating robust performance despite the low prevalence of neoplasia. The team attributes these gains to the multi-stage segmentation-focused pretraining pipeline and the use of high-resolution image inputs. Nevertheless, they acknowledge remaining challenges with false positives and mask granularity, suggesting opportunities for further refinement through improved localization or task-specific supervision.

The team also identified important drawbacks. Their reliance on heterogeneous pretraining datasets—spanning polyp and Barrett’s esophagus segmentation—may introduce domain mismatch in feature representations. More critically, they observed a substantial discrepancy between strong cross-validation metrics and weaker leaderboard performance, raising concerns about potential image overlap between the EVC datasets used in pretraining and the RARE25 challenge images. Since overlapping images could not be definitively identified, the team cautions that part of the observed performance gains may reflect unintended data leakage rather than true generalization.

\subsection{Team MEDAI}
Team MEDAI proposed a binary classification framework for early Barrett’s neoplasia detection built on a ResNet50 backbone pretrained on the domain-specific GastroNet-5M dataset. Their approach specifically targets the severe class imbalance present in RARE25 by incorporating Balanced MixUp, which interpolates both within-class and cross-class samples to enforce smoother decision boundaries and reduce overfitting to the abundant negative cases. To improve the reliability of probabilistic outputs, the method employs Platt calibration, fitting a logistic function to validation logits to mitigate confidence bias. Training stability is further enhanced through exponential moving average (EMA) updates, as well as AdamW optimization with warm-up and cosine scheduling.

The model operates on 224×224 inputs and uses only minimal augmentation beyond MixUp, reflecting a design that prioritizes calibration and class-balance rather than large-scale perturbations. While all RARE25 samples were used for training, the team used an external 100-image Early Barrett’s Cancer Detection dataset—unseen during training—as a dedicated validation and calibration set. The final system consists of a lightweight ResNet50 architecture (23.5M parameters), with Platt scaling adding only two additional learned scalars.

Across their ablation studies, the team reported clear and consistent gains from each methodological component. Starting with a GastroNet-initialized ResNet50 baseline, Balanced MixUp improved minority-class sensitivity and significantly increased AUROC and PPV@90\%Recall. Adding Platt calibration yielded the strongest overall results (AUROC 0.893, AUPRC 0.455), with improved calibration over raw sigmoid outputs and more stable precision under high-recall operating points. EMA contributed to more reliable convergence and reduced run-to-run variability.

The authors note several limitations. Since Platt scaling depends on the distribution of the validation set, its generalization may deteriorate under domain shift—a realistic challenge given the multicenter nature of RARE25. Balanced MixUp, although effective for imbalance mitigation, introduces synthetic interpolated samples that may not fully capture the subtle morphological diversity of early neoplasia, potentially blurring clinically meaningful patterns. Finally, the added training components (MixUp, EMA, calibration) increase pipeline complexity and require sensitive hyperparameter tuning, and Platt’s linear correction may be insufficient for cases with highly nonlinear miscalibration.

\subsection{Team AJLG ETRI}

Team AJLG ETRI introduced FocalScope. Their approach integrates a domain-specific preprocessing pipeline with a ConvNeXt-Base backbone~\citep{convnext} and an adaptive variant of Focal Loss tailored to clinical screening constraints. The preprocessing stage removes black borders via FOV masking, applies white-balance correction for inter-device consistency, and enhances mucosal contrast using CLAHE in LAB space—steps aimed at reducing acquisition variability across centres. For classification, the model relies on ImageNet-initialized ConvNeXt-Base with a simplified binary head and sigmoid output, optimized under a strong class-imbalance regime.

Training is fully supervised using the RARE25 dataset without external data. The team employs 5-fold stratified cross-validation, WeightedRandomSampling, and a clinically motivated weighting scheme in the Focal Loss (adaptive $\alpha$ and $\gamma=2$) to emphasize hard positive cases. Early stopping is driven directly by improvements in PPV@90\% recall, reflecting the intended real-world screening use case. At inference, the model uses a conservative test-time augmentation strategy—mild rotations and small brightness adjustments validated to preserve anatomical integrity—with outlier removal before averaging predictions.

Across cross-validation, FocalScope achieved a mean PPV@90\%recall of 0.2505. Results were consistent across folds, with most models converging between epochs 25–35, suggesting stable training dynamics despite limited positive data. The team attributes performance to the combination of endoscopy-specific preprocessing, ConvNeXt’s strong feature extraction capacity, and the adaptive weighting strategy focused on clinically meaningful recall-constrained PPV.

The authors highlight several limitations. The PPV remains relatively low, meaning the system still generates a substantial false-positive burden that could strain clinical workflows. The ConvNeXt-Base backbone (88M parameters) and test-time augmentation incur non-trivial computational overhead, which may hinder deployment in low-resource environments. The reliance on specific preprocessing assumptions—particularly FOV masking thresholds and CLAHE settings—may limit robustness to other endoscopic systems or imaging protocols. Additionally, the scarcity of positive cases restricts the model’s exposure to rare phenotypes, and the medically conservative TTA design may forgo augmentations that could improve generalization. Although optimized for Barrett’s neoplasia, transferring the pipeline to other endoscopic tasks would likely require substantial re-engineering.

\subsection{Team sk}
Team sk proposed a ConvNeXt-Base (V1) model~\citep{convnext} for classifying Barrett’s esophagus in endoscopic images, leveraging ImageNet-22k pretraining and a weighted Focal Loss to mitigate class imbalance. Their pipeline uses standard geometric augmentations—horizontal/vertical flips, 90° rotations, and shift/scale/rotate transforms—and trains ten independent models with different random seeds to improve robustness. During inference, predictions are averaged across all ten models, combined with 4-view rotational test-time augmentation, resulting in a 40-prediction ensemble per image. Input images are resized to 224×224 and normalized following standard ImageNet preprocessing.

Using a 50/50 random split (stratified by class), the ensemble achieved AUPRC 0.310, AUROC 0.832, and PPV@90\% recall 0.015 in the open development phase. On the full dataset evaluation, scores improved to AUPRC 0.527, AUROC 0.820, and PPV@90\% recall 0.163. The team notes that performance is sensitive to data splits and random seeds, and the reliance on heavy ensembling and rotational augmentations may limit scalability or real-time applicability.

\subsection{Team agaldran}
Team agaldran developed an ensemble-based approach using 25 ResNet50 models trained with logit-adjusted binary cross-entropy to address the extreme class imbalance of the RARE25 dataset. Multiple initialization strategies were explored, including ImageNet, GastroVision pretraining~\citep{GastroVision}, and GastroNet weights~\citep{BOERS2024103298}, with the latter providing the strongest performance. The model architecture consists of a ResNet50 backbone (25.6M parameters each), trained with N-Adam (lr=1e-4), cosine annealing, a batch size of 32, and light oversampling. All images were resized to 512×512 and intensity-normalized, and test-time augmentation was applied during inference to stabilize predictions. No additional external datasets were used beyond GastroNet’s published pretraining weights.

For evaluation, the dataset was split using five-fold stratified cross-validation (80/20 per fold), ensuring proportional representation of neoplasia cases. Across different initialization settings, the GastroNet-initialized ensemble achieved the strongest results, particularly under the challenge’s PPV@90\% recall metric. The authors note that internal cross-validation performance varied substantially across folds due to the very small number of positive samples, making PPV@90\% recall difficult to estimate reliably. Nonetheless, the final method demonstrated consistent improvements over ImageNet-initialized baselines and provided a competitive submission to the challenge.

\subsection{Team SAINTS}
This team employed a DeiT-Base model pre-trained on ImageNet-21k and adapted it for binary classification of neoplasia versus non-dysplastic Barrett’s esophagus. All images were resized to 224×224 and normalized using ImageNet statistics. Training used the default augmentation pipeline from the timm library, while validation relied solely on resizing and normalization. To compensate for the severe class imbalance, the authors applied a WeightedRandomSampler within an 80/20 stratified dataset split, ensuring that positive (neoplasia) samples were oversampled during training and were present proportionally in both training and validation sets. The final model contained approximately 86 million parameters, consistent with the DeiT-Base architecture.

On the 20\% held-out validation set, the model achieved an overall accuracy of 97.25\%, largely reflecting the dominance of negative cases. Class-balanced metrics provided deeper insight: precision reached 0.7778, recall 0.6562, and F1 score 0.7119. Under the clinically important PPV@90\%RECALL metric, performance was 0.0890 with a low optimal threshold (0.0078), illustrating the challenge of achieving high positive predictive value while maintaining strict sensitivity requirements.

\subsection{Team ReMIC}
The proposed method, CINDIE (Center-Invariant Neoplasia Detection under Extreme Class Imbalance via Domain-Adaptive Ensembles), tackles the dual challenge of severe class imbalance and multi-center variability in neoplasia detection. The system is built as an ensemble of four ImageNet-pretrained models: ConvNeXt v1, ConvNeXt v2, and two Hiera models. To ensure center-invariant learning, the ensemble incorporates domain-adaptation mechanisms such as gradient reversal layers. Supervised contrastive learning (“Tale of Two Classes”) reduces the over-dominance of the majority class, while cross-center mixup teaches the models to prioritize pathology-relevant features over dataset-specific artifacts.

The training pipeline integrates data from MICCAI 2015~\citep{EndoVIS_Barrett}, Kvasir~\citep{KVASIR}, EDD2020~\citep{EDD2020}, and GastroVision~\cite{GastroVision}, providing broader variability in non-dysplastic and neoplastic samples. A CLIP-based clustering strategy is used to prevent leakage when splitting data into 80\% training and 20\% validation, and problematic NDBE outliers are removed after inspection. Preprocessing includes synthetic data generation (SMOTE-like oversampling + FGSM), balanced batch sampling, and artificial corner masks to avoid shortcut learning. Models are trained at 224×224 resolution using AdamW and One-Cycle scheduling. Inference relies on ensemble averaging and extensive test-time augmentation, including crops, flips, rotations, color jitter, and Gaussian blur.

In the Open Development Phase, CINDIE shows strong discriminative performance (AUROC=0.78) but faces the expected precision drop under extreme 1\% prevalence (PPV@90\%RECALL=0.01), limiting reliability at very high recall. Although the model effectively ranks neoplastic cases, operating at stringent recall thresholds results in high false-positive rates. Main drawbacks include the computational overhead of a multi-model ensemble with heavy TTA and reduced interpretability due to the black-box nature of the combined architectures, potentially complicating real-world clinical adoption.

\subsection{Team MIRC}
The proposed method focuses on addressing the severe class imbalance and limited dataset size in Barrett’s neoplasia detection by applying the Asymmetric Loss (ASL) function, which downweights easy negatives and emphasizes difficult or potentially mislabeled samples. Built on a ResNet-50 backbone pretrained on ImageNet (23M parameters), the approach uses classical supervised learning with a hyperparameter grid search to optimize performance based on AUC and PPV@90\%RECALL. Training relies heavily on strong data augmentation—random cropping, resizing to 224×224, normalization, and random flips—to increase sample diversity, even at the risk of cropping out pathology. This integrates naturally with ASL, which is designed to handle mislabeled data. The dataset split uses the two centres of the open training set, assigning one center to training and the other to validation, resulting in a natural yet highly imbalanced distribution.

In development, the model selection strategy prioritized maximizing AUC while additionally constraining model behavior through PPV@90\%RECALL to ensure confident predictions. Although this simple pipeline demonstrates how far classical supervised methods can be pushed on a challenging task, it also carries drawbacks. The reliance on strong cropping-based augmentation may introduce systematic mislabeling, and the model’s performance can be sensitive to hyperparameter choices. Furthermore, the method does not incorporate more advanced imbalance-handling strategies, domain adaptation, or post-processing, which may limit its competitiveness compared to more complex approaches.

\subsection{Team iRARE}
The ICANSee proposed method enhances neoplasia detection in endoscopic images by leveraging a ResNet-50 backbone with GastroNet-5M domain-specific pretraining. The approach incorporates three key innovations: focal loss to emphasize learning on rare positive cases, split-aware oversampling to address severe class imbalance without data leakage, and multi-phase fine-tuning with differential learning rates. During training, the backbone is initially frozen for several epochs to allow the classification head to adapt, after which the entire network is fine-tuned. Input images are normalized using ImageNet statistics, with mild augmentations such as horizontal flips, small rotations, and subtle color jitter applied during training, while validation and test sets are only resized and normalized to maintain consistency.

The training strategy uses an 80/20 stratified split, with class weights computed from the training set and applied via a weighted sampler to ensure balanced exposure to rare neoplasia samples. Systematic hyperparameter tuning was performed across learning rates, focal loss $\alpha$ and $\gamma$ values, and backbone freeze epochs. The final model achieves robust validation performance, with an AUPRC of 0.81, demonstrating effective handling of class imbalance and strong discriminative ability while maintaining generalization on held-out data.

\subsection{Comparison of methods}
Ensembling emerged as a dominant methodological strategy across submissions. Several teams explicitly relied on large ensembles to mitigate variance and improve robustness, including IMSY (40-model ensemble), agaldran (25-model ensemble), and sk (10-model ensemble combined with rotational test-time augmentation). ReMIC and Jmees-inc adopted smaller, heterogeneous ensembles designed to capture architectural diversity rather than sheer model count. In contrast, the remaining teams submitted single-model pipelines (UT, MEDAI, AJLG\_ETRI, SAINTS, MIRC, and iRARE), typically compensating through tailored loss functions, data handling strategies, or preprocessing. Overall, ensemble usage reflected a trade-off between robustness and computational complexity that was handled differently across teams.

Pretraining strategy represented another major axis of methodological variation. Several teams leveraged domain-specific pretraining on gastrointestinal endoscopy data, including IMSY, MEDAI, agaldran, iRARE, and Jmees-inc, primarily using GastroNet-5M or related datasets. Other teams (UT, AJLG\_ETRI, sk, SAINTS, and MIRC) relied on generic large-scale visual pretraining such as ImageNet or ImageNet-21k. Jmees-inc uniquely employed a staged, segmentation-driven pretraining pipeline spanning polyp, Barrett’s esophagus, and cancer tasks, while IMSY combined domain-specialist CNNs with foundation-model fine-tuning via parameter-efficient adaptation (DINOv3 with LoRA). These choices highlight the variation of both task-aligned and general-purpose pretraining paradigms within the challenge.

Explicit post-hoc calibration was comparatively rare. MEDAI incorporated Platt scaling using an external validation dataset, and IMSY applied affine re-calibration to ensemble logits. In contrast, most teams addressed decision-threshold behavior indirectly through training-time mechanisms, including adaptive focal loss (AJLG\_ETRI), focal loss combined with oversampling (iRARE), asymmetric loss (MIRC), or logit-adjusted objectives (agaldran). This distinction reflects two prevailing methods: treating calibration as a separate post-processing step versus embedding operating-point control into the loss design. The limited adoption of explicit calibration suggests that probability alignment remains an underexplored component in low-prevalence screening pipelines.

Finally, internal validation strategies varied substantially, reflecting the challenges posed by the limited number of neoplasia samples. IMSY explored multiple validation schemes, including center-based splits and reserved internal test sets. UT, Jmees-inc, and agaldran employed stratified five-fold cross-validation, while MEDAI relied on an external dataset for validation and calibration. Other approaches included a 50/50 random split (sk), standard 80/20 splits (SAINTS and ReMIC, with ReMIC further incorporating CLIP-based clustering to reduce leakage risk), and a center-based split (MIRC). Together, these strategies illustrate the lack of a single dominant validation paradigm and underscore the methodological sensitivity of internal evaluation in extremely low-prevalence settings.

\section{Results}

\subsection{Test set results}

\begin{figure*}[t]
    \centering

    \begin{subfigure}[t]{0.32\textwidth}
        \centering
        \includegraphics[width=\linewidth]{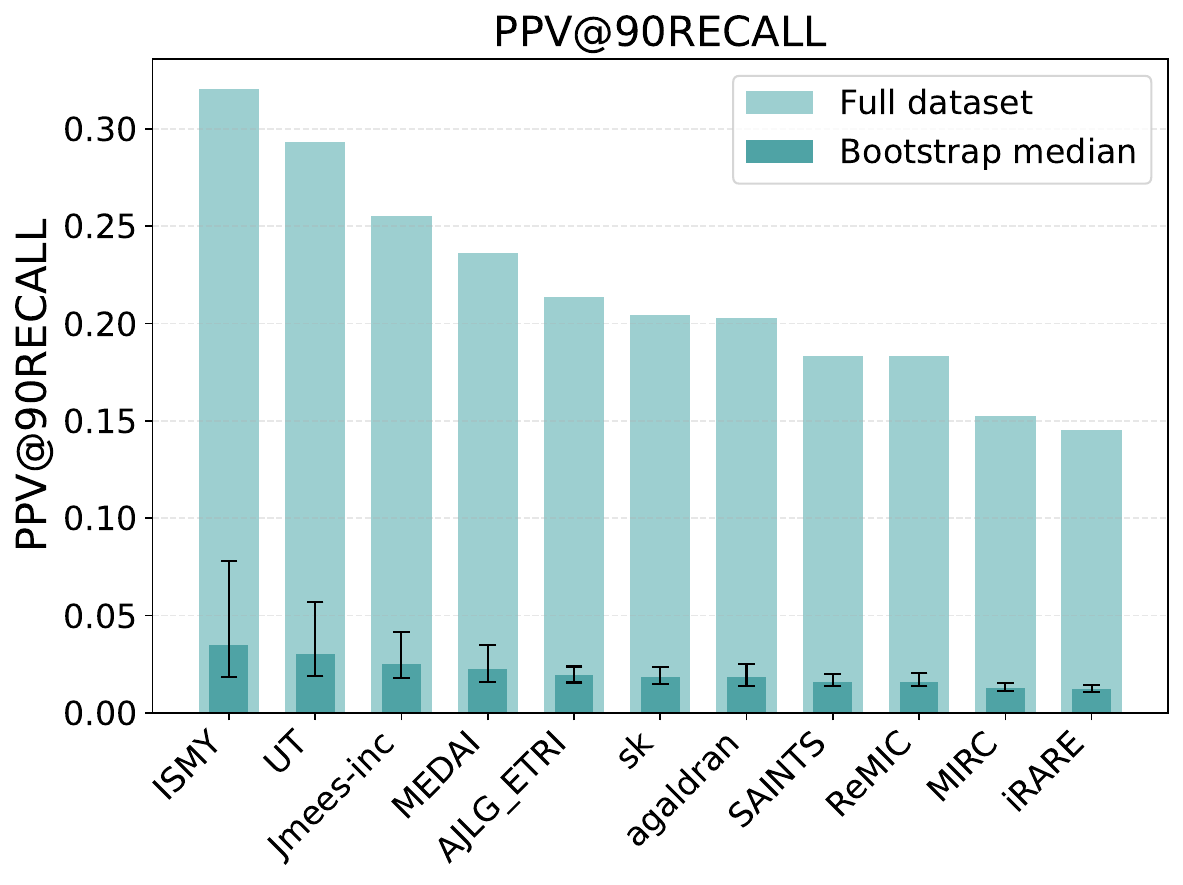}
        \caption{PPV@90\%RECALL}
        \label{fig:ppv90recal}
    \end{subfigure}
    \hfill
    \begin{subfigure}[t]{0.32\textwidth}
        \centering
        \includegraphics[width=\linewidth]{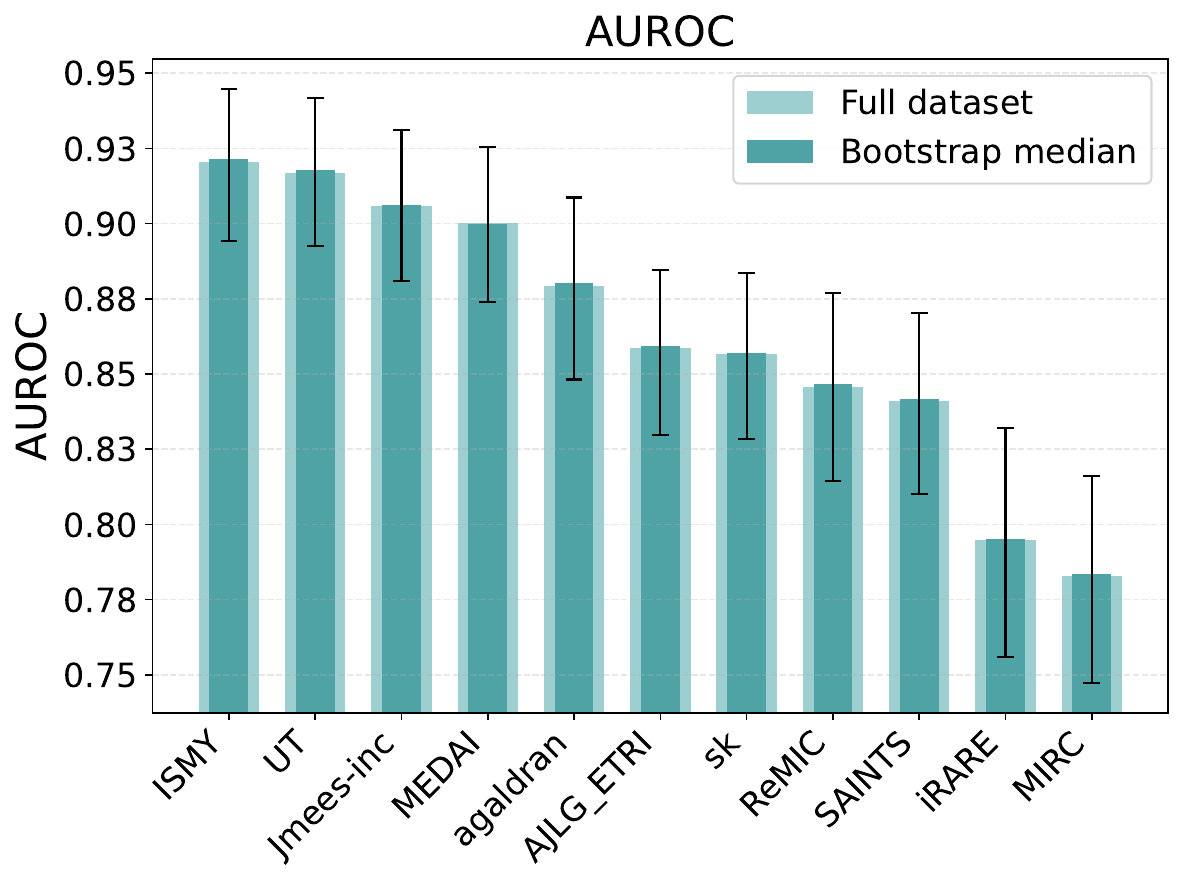}
        \caption{AUROC}
        \label{fig:auroc}
    \end{subfigure}
    \hfill
    \begin{subfigure}[t]{0.32\textwidth}
        \centering
        \includegraphics[width=\linewidth]{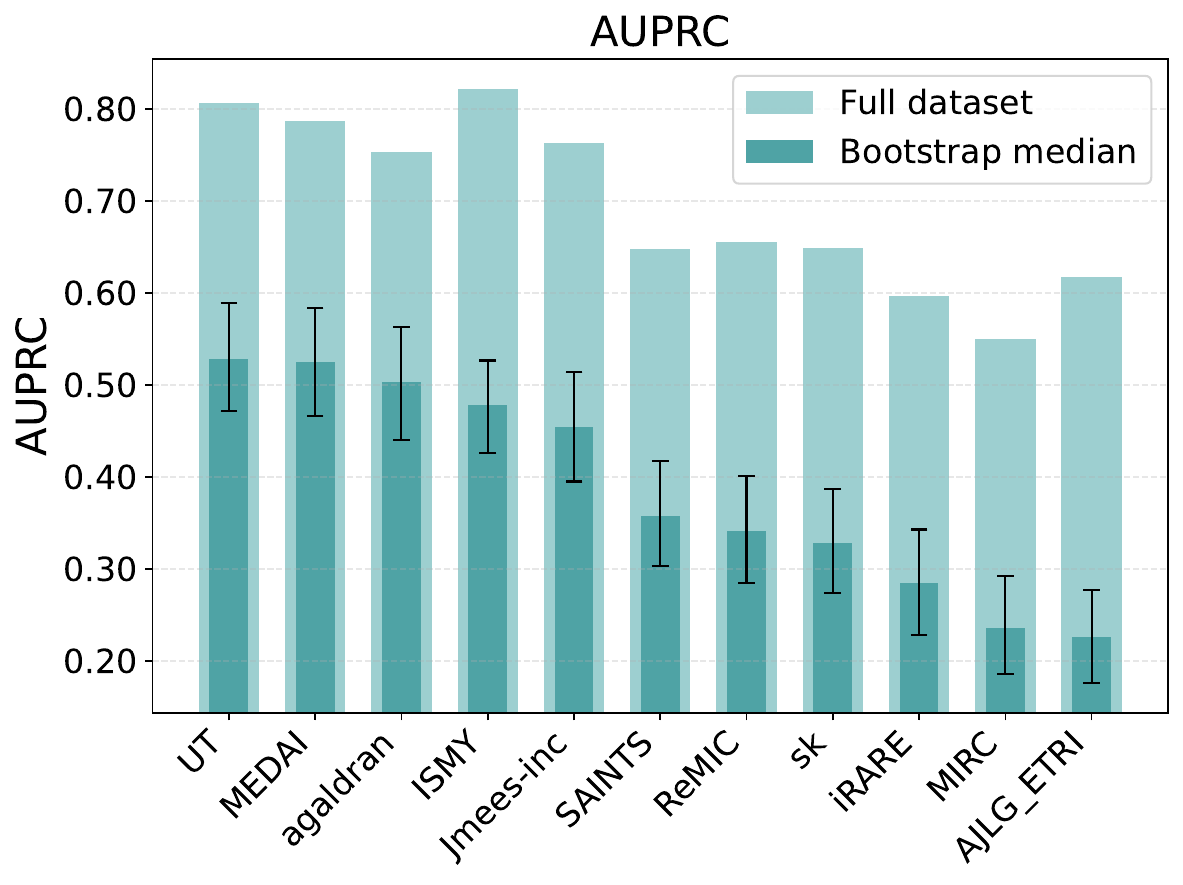}
        \caption{AUPRC}
        \label{fig:auprc}
    \end{subfigure}

    \caption{Results of all teams in the Closed Testing Phase. Performance is reported for (a) PPV@90\%RECALL, (b) AUROC, and (c) AUPRC. PPV@90\% recall was defined as the primary challenge metric and used to determine the final ranking, as it reflects performance at a clinically relevant operating point under real-world low-prevalence conditions.}
    \label{fig:results_subplots}
\end{figure*}

Figure~\ref{fig:results_subplots} summarizes the performance of all participating teams on the held-out test set. As shown in Fig.~\ref{fig:ppv90recal} and Fig.~\ref{fig:auroc}, Team IMSY achieved the strongest overall performance, attaining the highest median PPV@90\%RECALL across the prevalence-representative bootstrap samples (median PPV@90\%RECALL = 0.035). When evaluated on the full test dataset without resampling, the same method maintained the highest PPV@90\%RECALL, achieving a value of 0.320. Additionally, the IMSY Team achieved a AUROC of 0.920 on the full dataset.  In contrast, evaluation on the AUPRC exhibited a different ordering: Team UT achieved the highest median AUPRC across the bootstrapped samples (median AUPRC = 0.532; Fig.~\ref{fig:auprc}), whereas the highest AUPRC on the full test set was obtained by Team IMSY (AUPRC = 0.822), highlighting the sensitivity of precision–recall metrics to the faced prevalence. These results indicate that both top teams submitted methods with strong discriminative performance between neoplasia and non-dysplasia samples; however, the low PPV@90\%RECALL also highlights the large number of false positives when deployed in real-world settings.

Looking across teams, performance exhibits a clear stepwise progression rather than an early saturation. The top-ranked method achieves a 52\% improvement over the fifth-ranked method in terms of PPV@90\%RECALL evaluated on the full dataset, while the fifth-ranked method outperforms the tenth-ranked method by 50\%. This pattern indicates incremental gains across the leaderboard rather than a single dominant solution, as also becomes clear in Fig~\ref{fig:results_subplots}. A direct comparison between the two leading submissions (Team IMSY and Team UT) under the bootstrap evaluation further illustrates that the largest performance differences are observed (17\%) on the primary operating metric (PPV@90\%RECALL), with smaller (0.3\%) or inversely signed (-9\%) differences on ranking-oriented metrics such as AUPRC and AUROC. These observations highlight that improvements at strict high-recall operating points remain particularly challenging.

\begin{figure*}
    \centering
    \includegraphics[width=\textwidth]{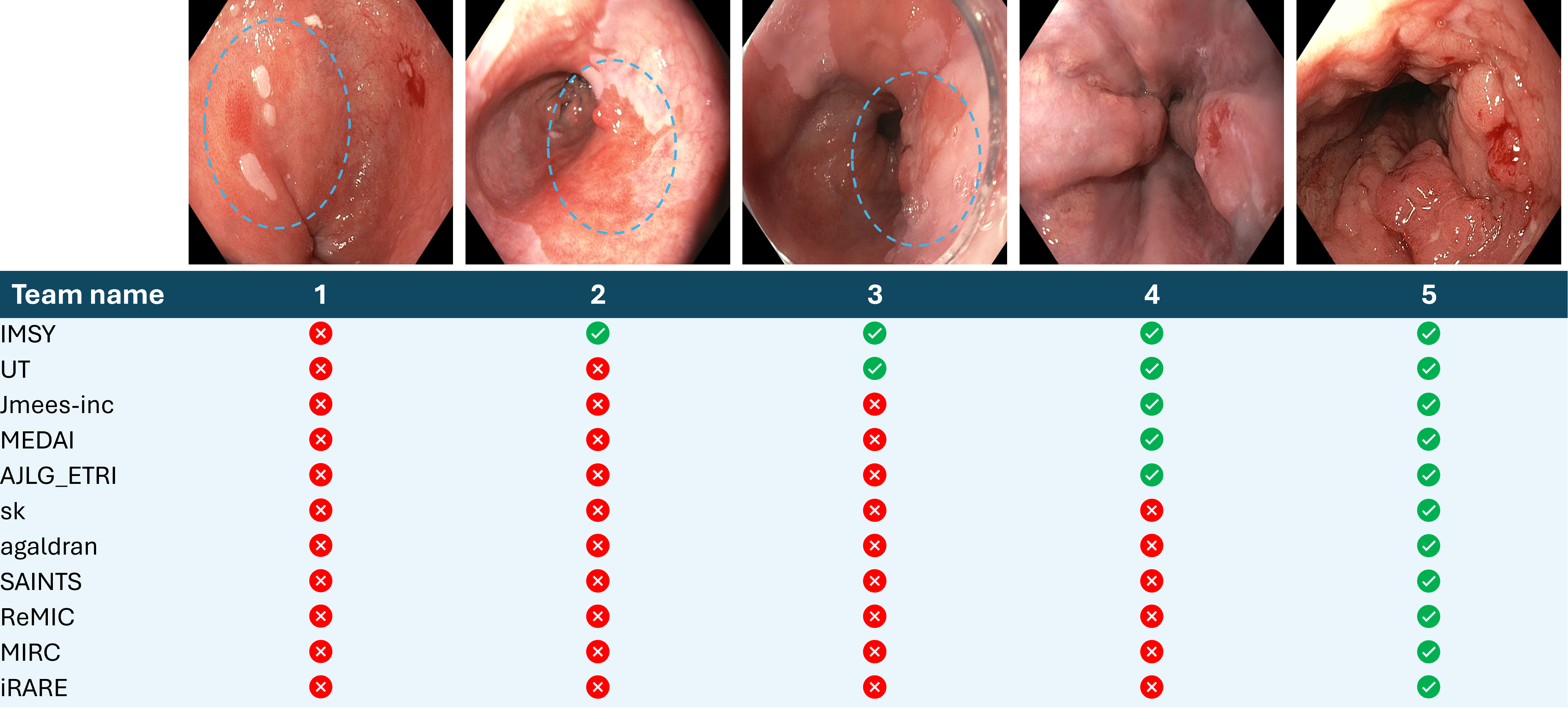}
    \caption{Representative endoscopic images containing neoplastic lesions, ordered from very subtle to severe appearance (left to right). In the first three examples, localized neoplastic regions are indicated by blue dotted outlines, whereas in the last two examples neoplasia involves a large portion of the visible mucosal surface. For each image, the classification outcome of the submitted methods is indicated, showing whether each team correctly classified the image.}

    \label{fig:visual examples}
\end{figure*}

Qualitative examples illustrating the behavior of different models are shown in Fig.~\ref{fig:visual examples}. These examples reveal that most methods struggle with very subtle manifestations of early neoplasia, particularly in cases where visual cues are faint or ambiguous (first and second examples). Nevertheless, the top-performing methods occasionally succeed in identifying such subtle lesions, suggesting effective utilization of weak discriminative signals. In contrast, more advanced neoplastic cases with clearer visual characteristics are consistently detected by all teams, underscoring that performance differences primarily emerge in the most clinically challenging early-stage cases.

\subsection{Ranking stability}
\begin{figure*}[t]
    \centering
    \begin{subfigure}[t]{0.32\textwidth}
        \centering
        \includegraphics[width=\linewidth]{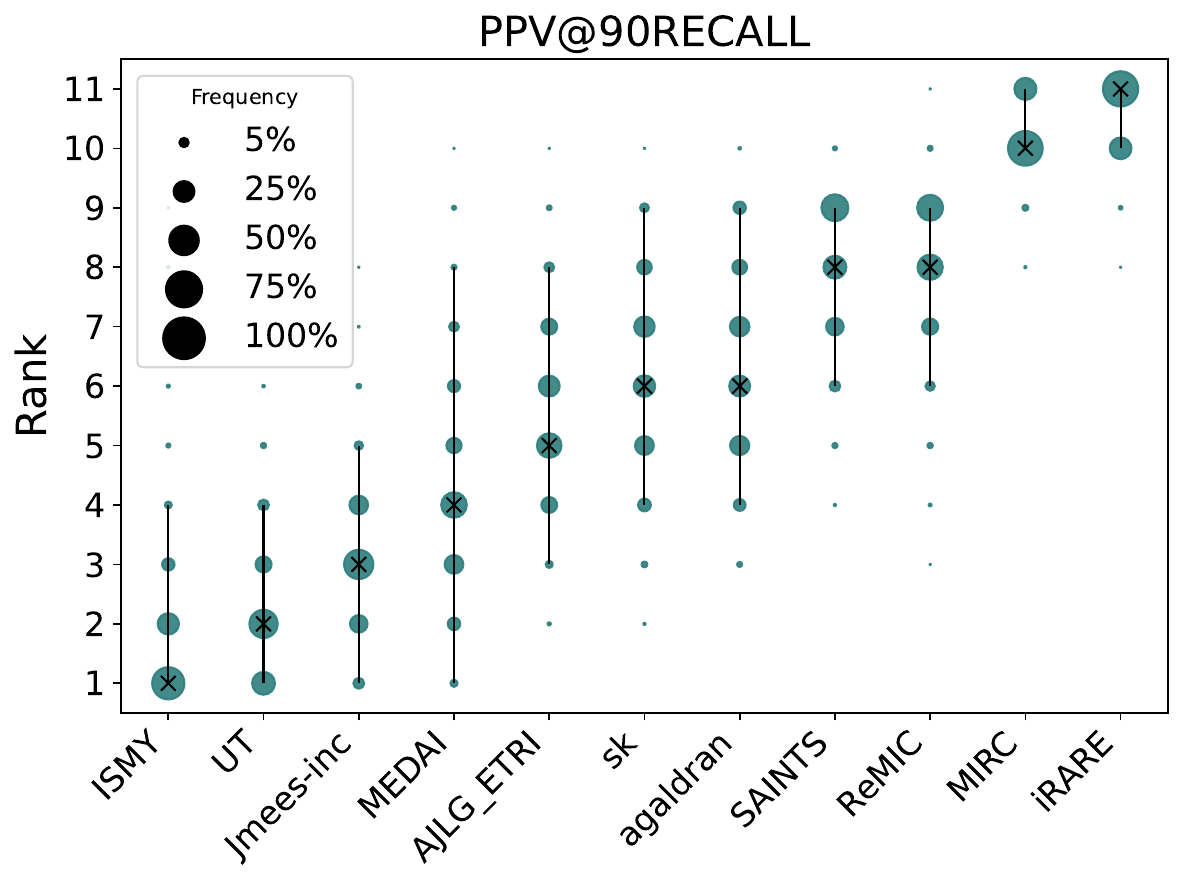}
        \caption{Ranking by PPV@90\%RECALL}
        \label{fig:rank_ppv90recal}
    \end{subfigure}
    \hfill
    \begin{subfigure}[t]{0.32\textwidth}
        \centering
        \includegraphics[width=\linewidth]{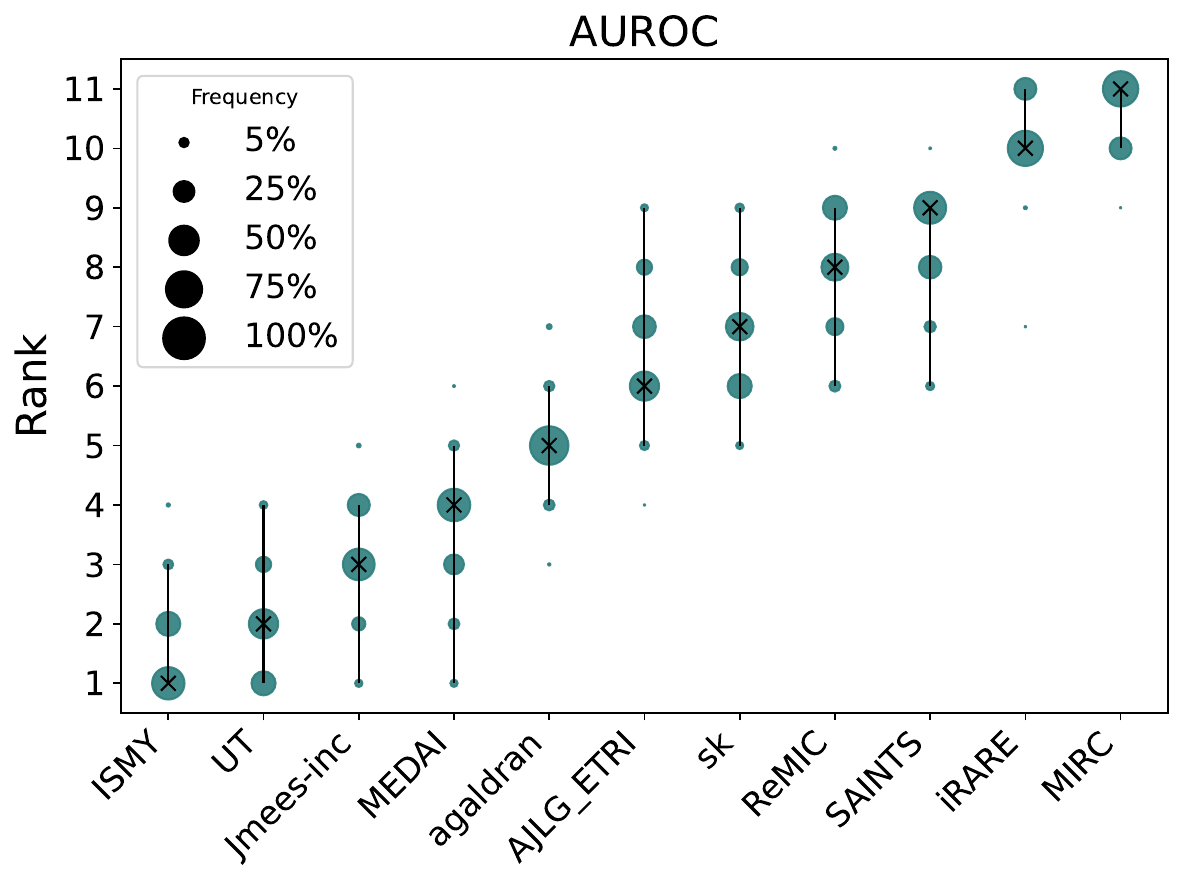}
        \caption{Ranking by AUROC}
        \label{fig:rank_auroc}
    \end{subfigure}
    \hfill
    \begin{subfigure}[t]{0.32\textwidth}
        \centering
        \includegraphics[width=\linewidth]{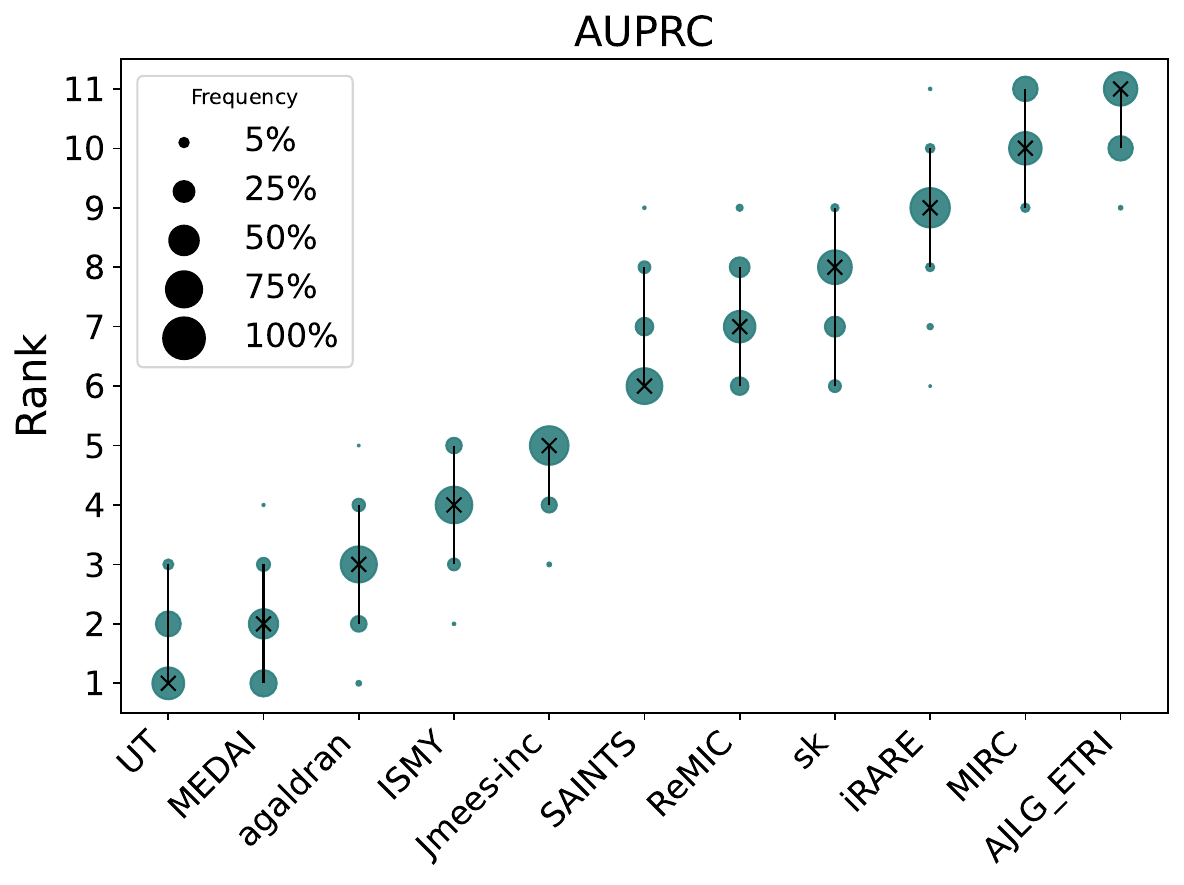}
        \caption{Ranking by AUPRC}
        \label{fig:rank_auprc}
    \end{subfigure}

    \caption{Team rankings in the Closed Testing Phase under different evaluation metrics. Each panel shows the relative ranking of teams when ordered by (a) PPV@90\%RECALL, (b) AUROC, and (c) AUPRC. Blob size is proportional to the frequency of the rank achieved based on bootstrapping (N = 1000) and the black lines indicate the 95\% confidence intervals.}
    \label{fig:ranking_subplots}
\end{figure*}

Fig.~\ref{fig:ranking_subplots} illustrates the variability in team rankings across the bootstrap test sets. Based on the PPV@90\%RECALL metric, team ISMY ranked first in 60.0\% of the bootstrap samples, followed by UT (30.0\%) and JMees-inc (6.9\%). A similar pattern was observed for AUROC, where ISMY achieved the first rank in 60.4\% of the bootstrap samples, followed by UT (33.5\%) and JMees-inc (3.1\%). 

For the AUPRC metric, the ranking distribution differed slightly: UT ranked first in 58.0\% of the bootstrap samples, followed by MEDAI (40.2\%) and agaldran (1.4\%). These results indicate that while the leading teams remain relatively consistent across bootstrap samples, some variability in the top positions is observed depending on the evaluation metric.

\section{Discussion}
The RARE25 challenge attracted broad international participation and produced a diverse set of methodological choices, underscoring that low-prevalence detection in routine surveillance remains both clinically important and technically unresolved. By evaluating methods on a large, prevalence-representative test set and using operating-point metrics designed for high sensitivity, RARE25 revealed concrete strengths and limitations of current CADe practice. Notably, top-performing systems combined domain-specific pretraining, foundation-model adaptation, and ensemble/post-hoc calibration strategies; nevertheless, positive predictive values at strict high-recall operating points remained below 4\% for all participating teams, illustrating the persistent false-positive burden of low-prevalence surveillance scenarios.

While several teams obtained strong ranking metrics, performance degraded substantially when methods were evaluated under realistic prevalence and uncertainty was considered. This gap highlights two important facts: (1)~internal cross-validation based on very few positives can be an unreliable indicator of real-world performance, and (2)~threshold-free ranking metrics (\emph{i.e.},~AUROC/AUPRC) can mask clinically important differences in absolute predictive values at deployment prevalence. For surveillance tasks, reporting operating-point metrics~(\emph{e.g.},~PPV at a clinically required recall) together with uncertainty estimates should be treated as \emph{essential} elements of method evaluation.

\subsection{Clinical impact}
The challenge outcomes indicate that modest but meaningful detection gains are attainable with current CADe techniques; however, the absolute PPVs observed at high-recall operating points remain low across many submissions, implying a substantial false-positive burden if systems are naively deployed at those thresholds. High false-positive rates can erode clinician trust, increase cognitive load, and prolong procedure time—outcomes that can offset any sensitivity gains. In short, successful clinical translation requires attention to both technical performance and practical workflow integration; reporting PPV at clinically relevant recalls and including clinician-centered endpoints in evaluations will accelerate responsible adoption.

\subsection{Limitations}
First, although the held-out test set was assembled to reflect real-world prevalence, the data originate from a limited set of devices (Exera III) and contributing centers; residual center-specific imaging characteristics and acquisition protocols may affect generalizability. Although some teams explicitly addressed center variability, fully characterizing and mitigating multicenter domain shift remains an open problem. Second, the number of positive examples available for supervised training is inevitably small for this problem, which increases the risk that cross-validation results are optimistic or unstable; this instability complicates both method development and fair model selection. Third, a subset of training images contained anonymization artifacts absent from the test set; although we documented this explicitly, such differences can create subtle distributional mismatches. 

\subsection{Future directions}
RARE25 highlights a clear methodological gap: despite the dominance of normal findings in surveillance workflows, no team submitted methods explicitly framed as anomaly or one-class detection. Prevalence-agnostic approaches may offer advantages for surveillance tasks where normal data are abundant and positive labels are scarce.

Prevalence-agnostic approaches, such as one-class classification, self-supervised representation learning on normal data, or hybrid models that combine anomaly scoring with limited supervised fine-tuning, may offer a more natural inductive bias for surveillance tasks. By learning a rich model of normal Barrett’s mucosa and identifying deviations as suspicious, such methods reduce dependence on scarce positive labels and may generalize more robustly under prevalence shift. Importantly, these approaches align with the clinical reality that new or rare neoplastic phenotypes may not be represented in historical training data.

Future benchmarks should therefore explicitly encourage or accommodate anomaly-detection–oriented submissions. Comparative studies that directly contrast supervised classifiers with prevalence-agnostic alternatives under identical evaluation protocols would be particularly valuable in clarifying when and how anomaly detection provides tangible advantages for CADe in surveillance settings.

Finally, integrating anomaly scores with calibrated decision rules and clinician-in-the-loop workflows may enable more flexible deployment strategies, where CADe systems function as conservative triage or attention-guidance tools rather than binary decision makers. Exploring such prevalence-agnostic paradigms represents a promising and largely unexplored direction for advancing CADe systems in low-prevalence clinical surveillance.

\section{Conclusion}
The RARE25 challenge introduces a prevalence-aware benchmark for computer-aided detection of Barrett’s esophagus neoplasia, addressing a central challenge in surveillance imaging where true positive findings are inherently rare. By evaluating methods under realistic prevalence conditions, the benchmark reveals limitations of commonly used development and evaluation practices that rely on artificially balanced datasets and aggregate performance metrics.

Across 11 participating teams, we observed substantial variability in model design and performance, particularly at clinically relevant high-recall operating points. Although several approaches achieved strong general discrimination, positive predictive values remained low for most submissions, highlighting the persistent false-positive burden that accompany low-prevalence detection tasks. The observed performance spread indicates that the benchmark is far from saturated and that meaningful gains remain possible through improved modeling, calibration, and evaluation strategies.

Beyond ranking submissions, RARE25 emphasizes broader methodological considerations for CADe in surveillance settings. Transparent evaluation under realistic prevalence, careful interpretation of operating-point–specific metrics, and robustness to prevalence shift are essential for clinical translation. The absence of prevalence-agnostic or anomaly-detection–based approaches among submissions further points to a promising direction for future research. By providing a large public dataset and a rigorous evaluation framework, RARE25 lays the foundation for the development of CADe systems that are better aligned with real-world surveillance workflows.

\bibliographystyle{model2-names.bst}\biboptions{authoryear}
\bibliography{refs}

\end{document}